\documentclass[10pt,twocolumn,letterpaper]{article}

\usepackage{iccv}              

\newcommand{\red}[1]{{\color{red}#1}}

\usepackage{amsmath}
\usepackage{amssymb}
\usepackage{booktabs}
\usepackage{multirow}
\usepackage{url}
\usepackage{bm}
\usepackage{diagbox}

\def\red#1{\textcolor{red}{\textbf{#1}}}
\def\blue#1{\textcolor{blue}{\textbf{#1}}}

\definecolor{iccvblue}{rgb}{0.21,0.49,0.74}
\usepackage[pagebackref,breaklinks,colorlinks,allcolors=iccvblue]{hyperref}

\def\paperID{8300} 
\def\confName{ICCV}
\def\confYear{2025}

\title{Dynamic Group Detection using VLM-augmented Temporal Groupness Graph}

\author{Kaname Yokoyama ~~ Chihiro Nakatani ~~ Norimichi Ukita\\
Toyota Technological Institute\\
{\tt\small \{sd25444,sd23501,ukita\}@toyota-ti.ac.jp}
}

\begin{document}
\maketitle

\begin{abstract}
This paper proposes dynamic human group detection in videos.
For detecting complex groups, not only the local appearance features of in-group members but also the global context of the scene
are important.
Such local and global appearance features in each frame are extracted using a Vision-Language Model (VLM) augmented for group detection in our method.
For further improvement, the group structure should be consistent over time.
While previous methods are stabilized on the assumption that groups are not changed in a video, our method detects dynamically changing groups by global optimization using a graph with all frames' groupness probabilities estimated by our groupness-augmented CLIP features.
Our experimental results demonstrate that our method outperforms state-of-the-art group detection methods on public datasets.
Code: \url{https://github.com/irajisamurai/VLM-GroupDetection.git}
\end{abstract}


\section{Introduction}
\label{section: introduction}

\begin{figure}[t]
  \begin{center}
  \includegraphics[width=\columnwidth]{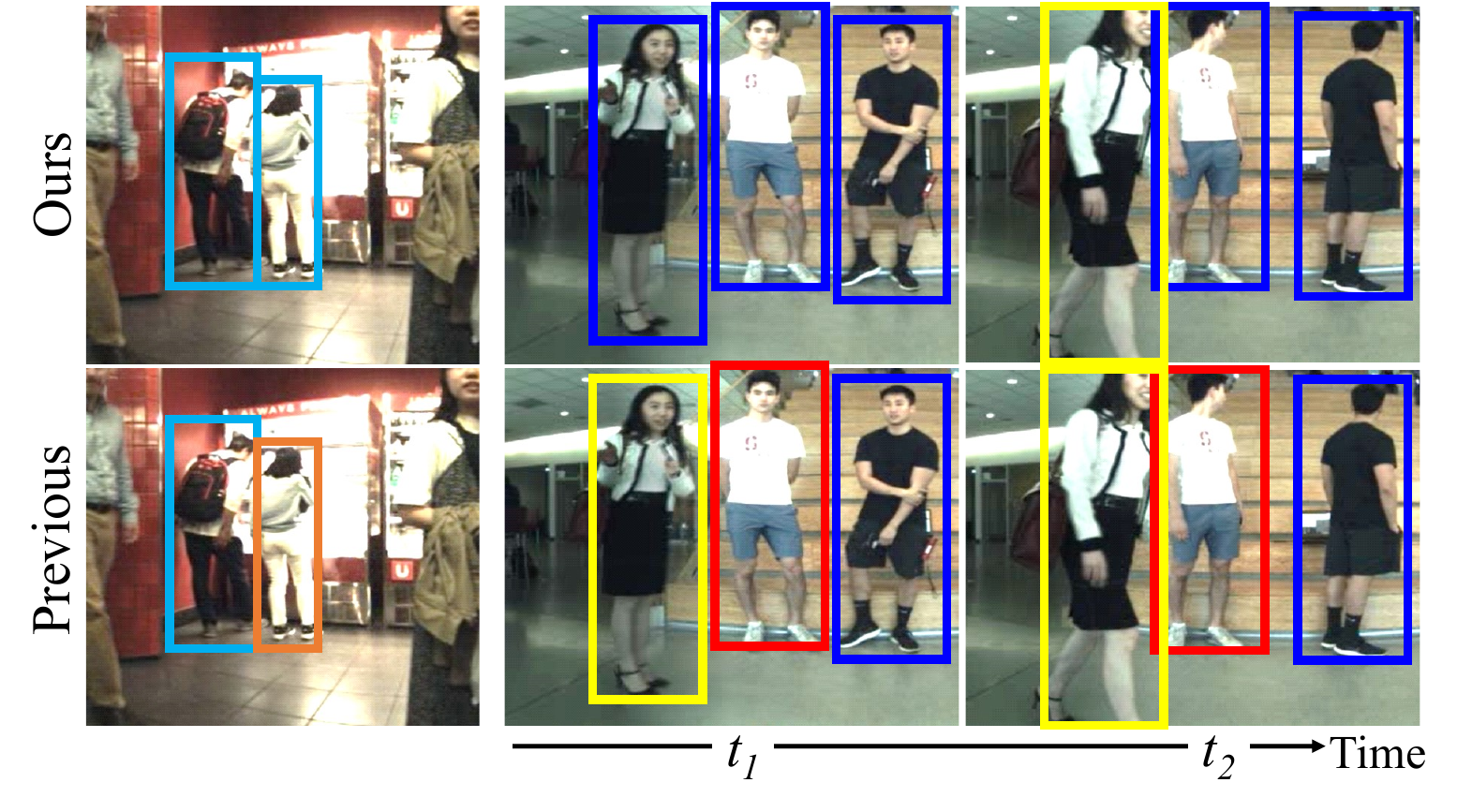}
  \end{center}
  \vspace*{-6mm}
  ~\hspace*{12mm}
  (a)
  ~\hspace*{30mm}
  (b)
  \caption{Effects of our method. People enclosed
  with the same color are detected as in-group members. (a) Our method
  using the spatial scene context 
  can detect two in-group members because they are not facing each other but browsing the same background object. (b) Our method can detect dynamic group structures so that a group of three members
  in $t_{1}$ splits into two groups in $t_{2}$.}
  \label{fig: teaser}
\end{figure}

To understand human social activities, 
the structure of human groups is a good cue because in-group members share a common purpose and/or behavior.
For example, the group structure is used for group activity recognition~\cite{groupdetection,groupdetection2,DBLP:conf/mva/NakataniSU21}, trajectory prediction~\cite{trajectoryprediction,trajectoryprediction2,trajectoryprediction3,DBLP:journals/cviu/UkitaMH16}, and anomaly detection~\cite{abnormaldetection}.
For group detection (e.g., Fig.~\ref{fig: teaser}), interactions among in-group members should be recognized.
While the dominant cue for recognizing such human interactions is the spatial relationship of people, various human attributes, such as face directions and postures, are also useful.

While the human attributes mentioned above can be recognized from the local appearance of each person, the spatial context observed in the image also provides helpful cues for group detection.
The usefulness of the spatial context is validated in various tasks, such as object detection~\cite{ObjectDetection,DBLP:journals/access/AkitaU23,ObjectDetection2,ObjectDetection3} and individual action recognition~\cite{HAR,actionclassificaction,actionclassificaction2}.
However, compared with these tasks that employ a relationship between {\em each person} and the scene, group detection is more difficult because it requires a complex relationship between {\em multiple people} and the scene
(e.g., Fig.~\ref{fig: teaser} (a)),
which is called {\bf spatial scene context}.

In addition to the spatial scene context, the {\bf temporal scene context} is useful.
For example, in the previous group detection methods~\cite{traj_gd,traj_gd2,traj_gd3,traj_gd5,PANDA,S3R2,GroupTrans}, group detection is stabilized under an assumption that the set of all groups is fixed (i.e., not changed) throughout a video. 
We call this group detection, in which only a single representative set of groups through a video is detected,
{\bf static group} detection.
However, people often go back and forth between different groups.
Static group detection can detect dynamically changing groups by dividing a video into short clips in each of which the group structure is not changed.
However, since we do not know the frames in which the group structure changes, it is impossible to divide a video into appropriate short clips by pre-processes; it is a chicken-and-egg problem.
Therefore, this paper aims at jointly achieving {\em stabilization of group detection} and {\em detection of dynamically changing groups} by employing the temporal scene context in a video.
We call this group detection, in which a set of temporal structures of dynamically changing groups is detected, {\bf dynamic group} detection
(e.g., Fig.~\ref{fig: teaser} (b)).


\begin{figure}[t]
  \begin{center}
    \includegraphics[width=\columnwidth]{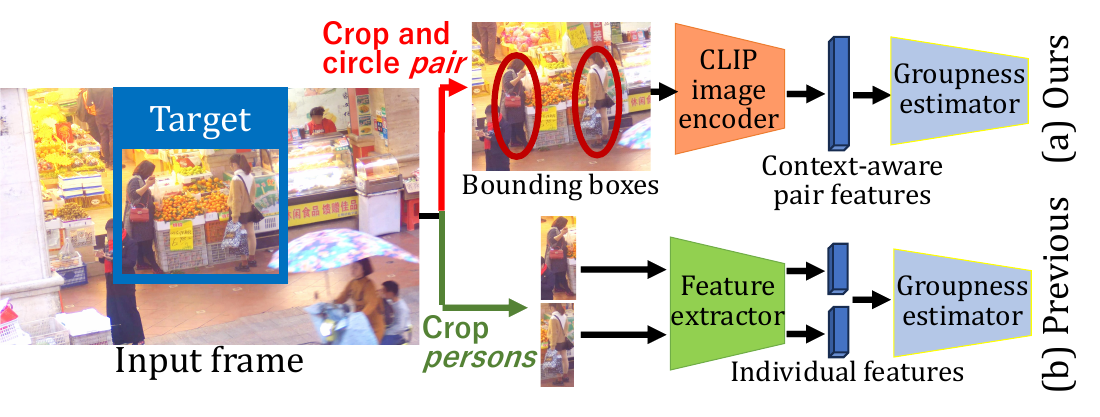}
  \end{center}
  \vspace{-6mm}
  \caption{Bounding boxes fed into the
  feature extractor. 
  (a) Ours: a bounding box includes a pair of circled people and the background. (b) Previous methods~\cite{S3R2,GroupTrans}: each person's bounding box is independently extracted.}
  \label{fig: image_features}
  \vspace*{-2mm}
\end{figure}

To learn the {\bf spatial scene context},
our method extracts image features from a bounding box, including people and the background (Fig.~\ref{fig: image_features} (a)), to gain two advantages compared to previous methods~\cite{GroupTrans,S3R2,jrdb}.
(1) In~\cite{GroupTrans,S3R2,jrdb}, the features of each person are extracted independently, then the extracted features of people are
compared and/or 
merged to estimate their groupness.
Our method directly estimates the groupness
from a bounding box including them.
(2) Not only the local features of people but also the spatial scene context, including the background, are used.
To understand a 
group structure based on the 
spatial scene context
from the bounding box mentioned above,
this paper proposes to augment a Vision-Language Model (VLM), such as CLIP~\cite{clip},
as visual features.

Such visual features are extracted independently in each frame of a video in our method, following previous methods~\cite{traj_gd,traj_gd2,traj_gd3,traj_gd5,PANDA,S3R2,GroupTrans}.
However, unlike the previous methods in which the detected groups of all frames are merged (e.g., averaged), our method temporally connects the features of all frames for dynamic group detection
to employ the {\bf temporal scene context}.

The contributions of this paper are as follows:
\begin{itemize}
\item  In each frame,
the {\bf spatial scene context} expressing
complex
interactions extracted by a VLM (i.e., CLIP) is used to estimate the so-called {\em groupness probability} to express whether or not each pair is in the same group.
Our augmented CLIP features
outperform
visual
feature extractors trained only with images, such as DINOv2~\cite{dinov2},
and improve the framewise group detection, resulting in better static and dynamic group detections.
\item Framewise groupness probabilities are temporally connected to construct a graph representing the {\bf temporal scene context}. This graph, called a {\bf temporal groupness graph} as contrasted with a {\bf static groupness graph} generated by merging all frames for static group detection, is divided into dynamic groups:
see Fig.~\ref{fig: graph_clustering}.
\item The results of dynamic group detection using our method can be easily converted to those of static group detection.
\item Our experiments on
public datasets
demonstrate that our method outperforms state-of-the-art
methods in various aspects of the static and dynamic detection tasks.
\end{itemize}


\section{Related Work}
\label{section: related}

\subsection{Group Detection}
\label{subsubsection: group}

In the group detection task, all observed people are divided into groups, including those each of which consists of only one person.
Unlike dense crowd tracking~\cite{DBLP:conf/eccv/ZhouTW12,DBLP:conf/aaai/LiCNW17,DBLP:journals/tmm/ZhouLDYYY24} in which particle-like crowd flows are represented as a set of feature points, group detection groups all individuals 

Early work on group detection~\cite{traj_gd2,traj_gd3} depends only on the spatial relationship (i.e., locations) among people.
In addition to the locations, facial directions are used to evaluate the so-called F-formation system~\cite{f-formation}, where groups are detected in conversational scenarios~\cite{f-formation_mathod,f-formation_mathod3,f-formation_mathod4}.
Group detection can be improved by using the locations and facial directions simultaneously~\cite{traj_gd}.

While the aforementioned methods formulate the groupness in a handcrafted fashion, such groupness formulations have recently been replaced by machine learning-based models.
In~\cite{traj_gd4,traj_gd5,DBLP:conf/eccv/PellegriniEG10,DBLP:conf/cvpr/YamaguchiBOB11,DBLP:journals/pami/SoleraCC16}, the characteristics of human trajectories are trained for group detection.
Other human attributes can also be merged, e.g., locations and facial directions~\cite{f-formation_mathod2,cnn_gd,lstm_gd,swofford2020improving} and locations and postures~\cite{S3R2}.
In these methods, all the human attributes are pre-processed.

\begin{figure*}[t]
  \begin{center}
    \includegraphics[width=\textwidth]{
    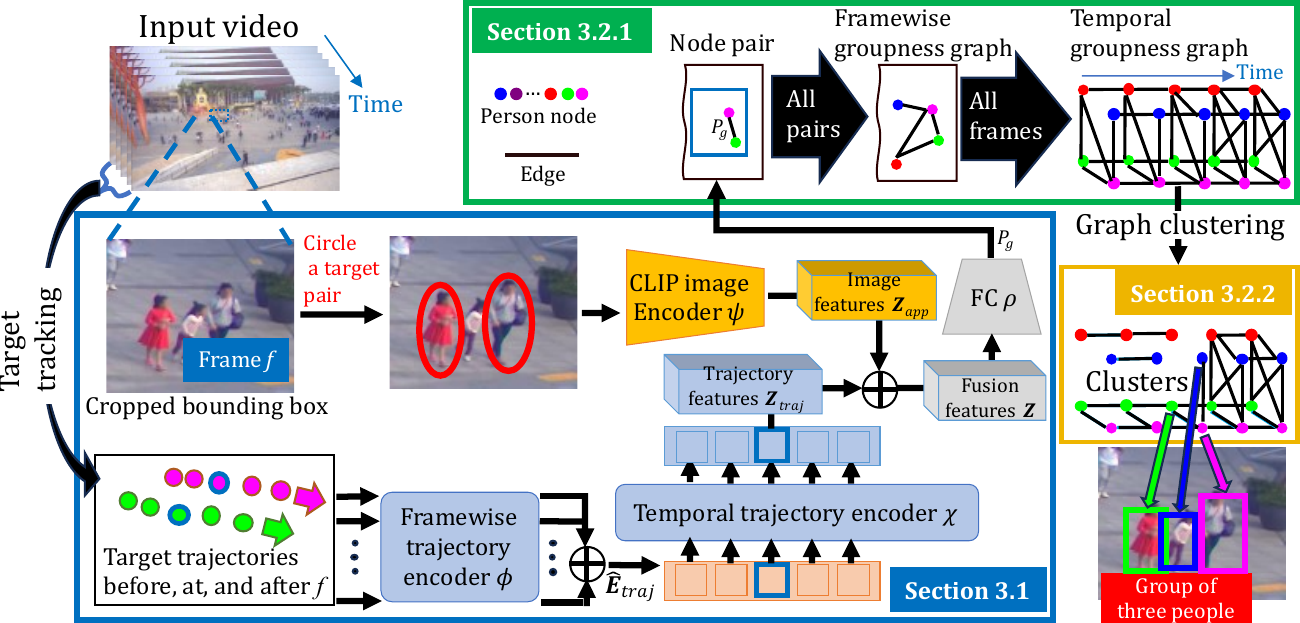}
  \end{center}
  \vspace{-2mm}
  \caption{Network architecture of our method. Extracted image and trajectory features are merged to construct the temporal groupness graph representing the probabilistic connectivity among people. A set of nodes in this graph is divided into clusters for group detection.}
  \label{fig: overview}
\end{figure*}

However, instead of the independent pre-processes, joint learning of the pre-processes and the final task benefits the final task 
in various tasks.
While such joint learning is also used for group detection, e.g., in still images~\cite{gnn_gd} and videos~\cite{groupdetection2,GroupTrans,jrdb}, these methods take human bounding boxes, in which important spatial context is unavailable.
Our work aims to extract useful group features from the spatial and temporal scene contexts represented by VLMs and the temporal groupness graph, respectively.

\subsection{Graph Clustering for Group Detection}
\label{subsection: clustering}

In recent group detection methods for videos~\cite{GroupTrans,S3R2,PANDA}, graph clustering divides a set of nodes in a graph into several clusters
as follows.
(1) The groupness probability of each person pair is estimated,
(2) a graph is constructed so that each node corresponds to each person, and
the nodes are connected via edges.
Each edge is given the estimated groupness probability between each pair,
and (3) clusters detected by graph clustering are regarded as groups.
These methods~\cite{GroupTrans,S3R2,PANDA} use spectral clustering~\cite{DBLP:conf/nips/NgJW01} and label propagation~\cite{label_propagation,lpa} as graph clustering.
While these simple clustering algorithms are useful for a small static groupness graph, dynamic group detection
is required to divide a large temporal groupness graph.
In addition, these simple algorithms need cluster-related na\"ive hyperparameters (e.g., the number of clusters and the max number of nodes in each cluster)
Unlike these simple algorithms, the Louvain algorithm~\cite{louvain} used in our method
efficiently allows for complex clustering
in such a large graph by maximizing the modularity of the graph without na\"ive hyperparameters.


\section{Dynamic Group Detection}
\label{section: method}

Figure~\ref{fig: overview} shows our
group detection network.
The groupness probability for each pair is estimated in each frame by our augmented CLIP image features with trajectory-based features (Sec.~\ref{subsection: pairwise_group_detection}).
All groupness probabilities are pairwise and framewise.
The
groupness probabilities in all frames in a video are used to construct the temporal groupness graph (Sec.~\ref{subsubsection: graph_contruction}) for dynamic group detection (Sec.~\ref{subsubsection: graph_clustering}).

\subsection{Groupness Probability Estimation}
\label{subsection: pairwise_group_detection}

The groupness probability 
is estimated from two features, i.e., trajectory and image features.


\subsubsection{Trajectory Feature Extraction}
\label{subsubsection: traj_extract}

The trajectory features are computed with two steps:

\noindent
{\bf 1st step (framewise feature extraction):}
Following~\cite{GroupTrans}, for any pair of people, 
a vector consisting of their appearance attributes such as bounding box locations and face directions (denoted by $\textbf{X}_{traj} \in \mathbb{R}^{D_{x}}$) is embedded to its latent variable, $\textbf{E}_{traj} \in \mathbb{R}^{D_{e}}$, by the framewise trajectory encoder, $\phi$, as follows:
\begin{equation}
    \textbf{E}_{traj} = \phi(\textbf{X}_{traj})
    \label{eq: embedding}
\end{equation} 

\noindent
{\bf 2nd step (temporal feature extraction):}
$\textbf{E}_{traj}$ of $T$ frames, which are selected before, in, and after frame $f$, are concatenated.
The concatenated features, $\hat{\textbf{E}}_{traj}$, are fed into the Transformer-based temporal trajectory encoder~\cite{transformer}, $\chi$, to obtain the trajectory features at frame $f$:
\begin{equation}
    \textbf{Z}_{traj} = \chi(\hat{\textbf{E}}_{traj}) \in \mathbb{R}^{D_{t}}
    \label{eq: trajectory_features}
\end{equation}


\subsubsection{Image Feature Extraction}
\label{subsubsection: image_extract}

\begin{figure}[t]
  \begin{center}
  \includegraphics[width=\columnwidth]{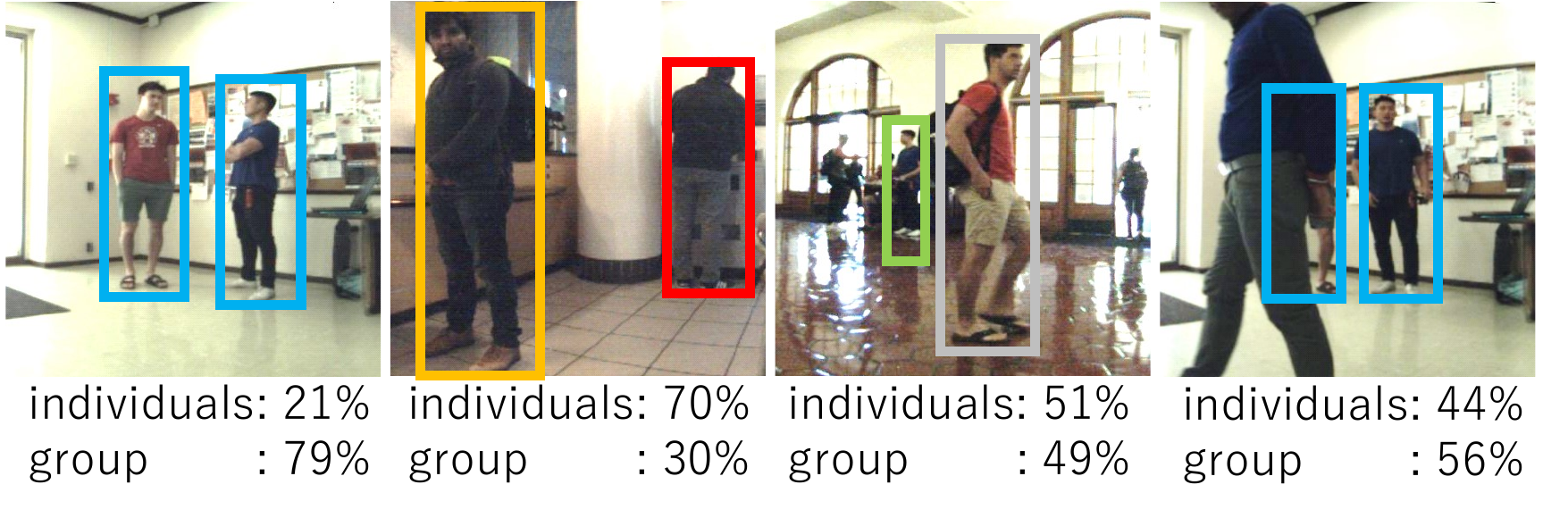}
  \end{center}
  \vspace*{-5mm}
  \hspace*{8mm}
  (a)
  \hspace*{15mm}
  (b)
  \hspace*{15mm}
  (c)
  \hspace*{15mm}
  (d)
  \vspace{0mm}
  \caption{Preliminary experiments with the pretrained CLIP for zero-shot group classification. While a bounding box with no visual prompt is fed into the image encoder,
  rectangles with the same color in each image enclose people in the same group of ground truth for visualization.
  }
  \label{fig: preliminary}
\end{figure}

\noindent
{\bf Preliminary experiments:}
We expect that CLIP is pretrained to understand 
what a human group is.
The potential of CLIP is validated in our experiments (Fig.~\ref{fig: preliminary}).
In these experiments, a bounding box including a target pair and the background is fed into the CLIP image encoder, $\psi$, to obtain its visual feature vector, $f_{v}$.
Two texts, ``individual people'' and ``a group of people''\footnote{These two texts are empirically chosen among similar text prompts.}, are fed into the CLIP text encoder independently to obtain their text feature vectors, $f_{i}$ and $f_{g}$.
The cosine similarities between $f_{v}$ and the text feature vectors are computed
and fed into the binary softmax function to obtain the probabilities that the target pair is in different groups and the same group.
In Fig.~\ref{fig: preliminary} (a) and (b), where two in-group members and two individual people are observed, respectively, the estimated probabilities are relatively good.
However, as shown in Fig.~\ref{fig: preliminary} (c), it is not easy for CLIP to pay attention to any pair because three or more people are observed.
Figure~\ref{fig: preliminary} (d) shows another difficulty. Since the left person enclosed by the box is occluded, it is difficult for CLIP to recognize whether or not the two people enclosed by boxes are in the same group.

\noindent
{\bf CLIP attention to a target pair:}
Even if three or more people are observed in a bounding box (e.g., Fig.~\ref{fig: preliminary} (c)),
our method proposes to direct the CLIP's attention to a target pair by circling each person of this pair (Fig.~\ref{fig: image_features} (a)).
The effectiveness of circling a region of interest for CLIP is validated
for the simple image-text classification task~\cite{redcircle}.
While only one region
is circled in each image in~\cite{redcircle}, our method applies this circling scheme to the group detection task so that two people are circled separately (Fig.~\ref{fig: image_features} (a)).

\noindent
{\bf Fine-tuning for groupness-augmented CLIP:}
While the potential of the pretrained CLIP is validated in our preliminary experiments, it is further finetuned in our method.
The objectives of this finetuning are twofold.
(1) CLIP is finetuned
in a supervised manner using the aforementioned red-circled pair. (2) In addition to the two classes used in our preliminary experiments (i.e., ``individual people'' and ``a group of people''), the ``occlusion'' class is added to cope with difficulty in occlusion handling, which is shown in Fig.~\ref{fig: preliminary} (d).
We call the features obtained by this finetuning the {\em groupness-augmented CLIP (GA-CLIP)} features.

Our method finetunes the CLIP image encoder, $\psi$, with the last softmax layer.
This finetuning is done with the image-text training data for three-class classification:
\begin{description}
    \item[Image:] 
    A cropped bounding box with a pair of two circled people is used, as described before.
    \item[Text:] Each 
    bounding box is labeled by either of
    ``a group of people,'' ``individual people,'' or ``occlusion.''
    This occlusion labels are available as annotations in the JRDB dataset~\cite{jrdb} used in this experiment. 
\end{description}
Our method uses ``occlusion'' to explicitly find that image features may be unreliable if the features are extracted from 
a bounding box in which at least either of the two circled people is occluded.
If image features are unreliable,
our method relies more on the trajectory features in the image-trajectory fusion features described in Sec.~\ref{subsubsection: fusion}.
    
Note that, in the following processes, only the GA-CLIP image encoder is used, but the last softmax layer is not used.

\noindent
{\bf Image feature extraction:}
While our trajectory features in frame $f$ are computed from $T$ frames, image features
in frame $f$ (denoted by $\textbf{Z}_{app}$) are extracted only from frame $f$:
\begin{equation}
    \textbf{Z}_{app} = \psi(\textbf{I}_{app}) \in \mathbb{R}^{D_{a}},
    \label{eq: image_features}
\end{equation}
where $\textbf{I}_{app} \in \mathbb{R}^{H \times W}$ denotes a cropped bounding box consisting of $H \times W$ pixels in frame $f$.


\subsubsection{Image-Trajectory Fusion Features for Groupness Probability Estimation}
\label{subsubsection: fusion}

The trajectory features, $\textbf{Z}_{traj}$, and the image features, $\textbf{Z}_{app}$, of a person pair are fused as $\textbf{Z} = \textbf{Z}_{app} \oplus \textbf{Z}_{traj} \in \mathbb{R}^{D_{a}+D_{t}}$, where $\oplus$ denotes a concatenation operator.
$\textbf{Z}$ is used for estimating the groupness probability between this pair (denoted by $\textbf{R}$) in each frame:
\begin{eqnarray}
        \textbf{R} &=& (P_{i}, P_{g})^{T} = SM(\rho(\textbf{Z})) \in \mathbb{R}^{2}, \label{eq: R}        ,
\end{eqnarray}
where $\rho$ and $SM$ denote fully connected layers and a softmax function, respectively. 
$P_{i}$ and $P_{g}$ are the probabilities that the pair are individual people and in the same group, respectively.
$\textbf{R}$ is estimated for all possible pairs of people in each frame.


\subsubsection{Joint Learning}
\label{subsubsection: joint}

With image-trajectory fusion features mentioned above, our method trains the three learnable networks, namely 
the trajectory feature extractors (i.e., $\phi$ in Eq.~(\ref{eq: embedding}) and $\chi$ in Eq.~(\ref{eq: trajectory_features})) and the image-trajectory fusion feature extractor (i.e., $\rho$ in Eq.~(\ref{eq: R})), while the GA-CLIP image encoder is fixed.
All of $\phi$, $\chi$, and $\rho$ are trained in a supervised joint learning manner.
In this joint learning, $\textbf{R}$ in Eq.~(\ref{eq: R}) is used to compute the cross entropy loss with its ground truth, $\textbf{R}_{gt} \in \{ (1, 0)^T, (0, 1)^T \}$.


\subsection{Dynamic Group Detection}
\label{subsection: graph_clustering}

\begin{figure}[t]
  \begin{center}
    \includegraphics[width=\columnwidth]{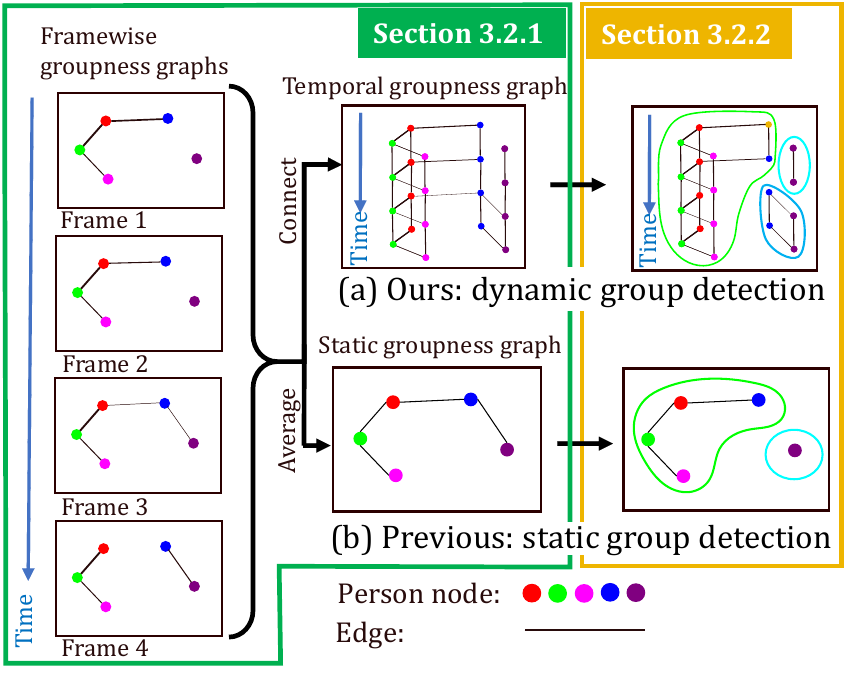}
  \end{center}
  \vspace{-3mm}
  \caption{Graph construction and clustering. (a) Ours: dynamic group detection by clustering the temporal groupness graph. (b) Previous methods: static group detection with a static groupness graph, which is generated by aggregating the graphs of all frames.}
  \label{fig: graph_clustering}
\end{figure}

\subsubsection{Graph Construction}
\label{subsubsection: graph_contruction}

Our method uses $P_{g}$ in Eq.~(\ref{eq: R}) to construct a graph for dynamic group detection in a video, as shown in the left-hand part of Fig.~\ref{fig: graph_clustering}, via the following two steps.

\noindent
{\bf Step 1:}
In each frame, a framewise groupness graph is constructed so that each detected person serves as a node.
$P_{g}$ of each pair is given to the edge that connects the nodes of this pair as its weight.

\noindent
{\bf Step 2:}
The framewise groupness graphs of all frames in a video are connected between subsequent frames to construct a temporal groupness graph.
Any
pair of nodes between frames $f$ and $f+1$ can be connected via an edge if this
temporal
pair's identification probability (denoted by $P_{t}$) is available and given to this edge. $P_{t}$ may be estimated by multiple object tracking (i.e., $0 \leq P_{t} \leq 1$) or annotated as
ground-truth
tracking IDs (i.e., $P_{t} = \{ 0, 1 \}$).

\subsubsection{Graph Clustering}
\label{subsubsection: graph_clustering}

All nodes in the temporal groupness graph are divided into clusters by graph clustering, as shown in the right-hand of Fig.~\ref{fig: graph_clustering}.
The clusters are regarded as dynamic groups.
For example,
in $f$-th framewise graph, nodes in the same cluster correspond to in-group members in frame $f$.
If in-group members differ between $f$ and $f+1$, that means the group
changes between $f$ and $f+1$.
This graph clustering is achieved by the Louvain~\cite{louvain}, as descrobed in Sec.~\ref{subsection: clustering}.


\section{Experimental Results}
\label{section: experiments}

\subsection{Datasets, Tasks, and Evaluation}
\label{subsection: exp_dataset}

\noindent
{\bf Datasets.}
The following two public datasets\footnote{PANDA dataset~\cite{PANDA}, which is another standard dataset, has only static group annotations. Therefore, experimental results on PANDA for static group detection are shown in the Supp.} are used:

{\bf JRDB dataset}~\cite{jrdb} includes 27 indoor and outdoor videos of first-person-view captured 
from a mobile robot.
Following~\cite{jrdb,S3R2,GroupTrans}, 20 and 7 videos are used for training and test, respectively.
2fps videos are sampled from the original videos.
In total, 1,419 and 404 frames are included in the sampled training and test videos, respectively.
Framewise group structures are annotated.

{\bf Caf\'{e} dataset}~\cite{DBLP:conf/eccv/KimSCK24} consists of 24 bird's-eye videos of
cafes.
Each video is split into short clips at six-second intervals.
Groups are annotated to each clip.
In our experiments, five clips are
concatenated to produce a 30-second video so that several 30-second clips capture dynamic group structure changes.
All clips are split into training, validation, and test splits in two different ways, split-by-view and split-by-place, which are called Caf\'{e}V and Caf\'{e}P, respectively.
The main paper shows the mean of the results of Caf\'{e}V and Caf\'{e}P, which is called Caf\'{e}.
The separate results of Caf\'{e}V and Caf\'{e}P are shown in the Supp.
As with JRDB, 2fps clips are sampled from the original clips.
In total, Caf\'{e}V and Caf\'{e}P have 60,135 and 78,871 training, 30,116 and 16,956 validation, and 30,078 and 24,502 test frames, respectively.
Unlike JRDB, Caf\'{e} has no annotations of occlusion for each person. In GA-CLIP training, therefore, each pair is labeled either ``a group of people'' or ``individual people.''


\noindent
{\bf Tasks.} Static and dynamic group detections are evaluated.

{\bf Static group detection} is tested to validate the generalizability (i.e., applicability to static detection) of our method.
All previous methods~\cite{groupdetection2,jrdb,dismat,S3R2,arg,GroupTrans,PANDA} in our experiments are static detection methods.
While our method is designed for dynamic detection, its framewise results are merged for static detection as follows.
The number of frames where each detected group is observed is counted to sort all groups in descending order.
In this order, the groups are selected one by one.
This group selection ends when all people are included in any of the selected groups.

{\bf Dynamic group detection} is the main task.
Three SoTA static group detection methods, S3R2~\cite{S3R2}, P-HAR~\cite{DBLP:conf/eccv/HanYLWFW22}, and GroupTransformer~\cite{GroupTrans}, are modified
for dynamic
detection,
following our method. That is, framewise graphs are not merged (e.g., averaged) but connected to construct a temporal groupness graph to divide it by graph clustering.
%
All those methods~\cite{S3R2,GroupTrans}, are provided by the authors' code.
As graph clustering, label propagation (LP)~\cite{lpa}, Clauset-Newman-Moore greedy modularity maximization (CNM)~\cite{CNM}, and Louvain algorithm~\cite{louvain} are tested.


\noindent
{\bf Evaluation.}
Following~\cite{PANDA,S3R2,GroupTrans}, the annotated human bounding boxes and temporal person IDs are used at training and test time, and detected groups are evaluated by the following criteria.
Given the sets of person IDs in detected and ground truth groups denoted by $G_{det}$ and $G_{gt}$, respectively.
If $\frac{ \left|G_{det}\cap G_{gt}\right|}{{\rm max}(\left|G_{det}\right|,\left|G_{gt}\right|)}>0.5$,
$G_{det}$ is regarded as a true-positive detection.
Otherwise, 
false-positive.
With these criteria, precision, recall, and F1 scores are obtained.


\subsection{Details}
\label{subsection: exp_details}

\subsubsection{Parameters}
\label{subsubsection: exp_parameteres}

\noindent
{\bf Architecture:}
Our method consists of 346M parameters (3091MiB), including 342M parameters of CLIP.
Louvain needs 476 and 917 MiB
on JRDB and Caf\'{e}, respectively.

\noindent
{\bf Hyper Parameters:}
The feature dimensions are
$D_{x} = 8$,
$D_{e} = 128$,
$D_{t} = 128 \times T$, and
$D_{a} = 768$.
All frames are used in each video (i.e., between 28 and 115 frames on JRDB and around 60 frames on Caf\'{e}).
As the CLIP image encoder,
our method uses ViT-L/14@336px that takes a bounding box consisting of $H \times W = 336 \times 336$  pixels.

\noindent {\bf CLIP finetuning:}
Adam is used with $\beta=(0.9, 0.99)$, $\epsilon =$ 1e-6, and ${\rm weight}\_{\rm decay} =$ 0.05.
The learning rate increases from 1e-8 to 2e-6 during the first 2 epochs, then decreases to 5e-8 with a cosine scheduler until 30 epochs.

\noindent {\bf Joint learning:}
While the weights of $\rho$ in Eq.~(\ref{eq: R}), $\phi$ in Eq.~(\ref{eq: embedding}), and $\chi$ in Eq.~(\ref{eq: trajectory_features}) are trained with the cross-entropy loss, as described in Sec.~\ref{subsubsection: joint}, for 30 epochs.
SGD is used.
In addition, ${\rm weight}\_{\rm decay} =$ 5e-4 and momentum=0.9 on PANDA.
The learning rate is fixed to 1e-4.


\subsection{Comparison with SoTA methods}
\label{subsection: exp_comparison}

\begin{table}[t]
  \centering
  \caption{Quantitative results of static group detection on JRDB.
  The best and second-best scores in each metric on each dataset are colored \red{red} and \blue{blue}, respectively 
  (also in Tables \ref{table: sota_comparison_static_cafe}, \ref{table: sota_comparison_dynamic_jrdb}, and \ref{table: sota_comparison_dynamic_cafe}).
  }
  \label{table: sota_comparison_static_JRDB}
    \vspace*{-1.5mm}
   {\footnotesize
   \begin{tabular}{l|ccc}
    \hline
   Method & Precision & Recall & F1 \\
    \hline\hline
    JRDB-Group~\cite{jrdb}  & 0.390 & 0.379 & 0.384 \\
    Dis.Mat+~\cite{dismat}   & 0.573 & 0.235 & 0.334\\
    GNN w/ GRU~\cite{S3R2}  & 0.434 & 0.286 & 0.345 \\
    ARG~\cite{arg}  &  0.325 & 0.384 & 0.352\\
    S3R2~\cite{S3R2}  &  0.577 & 0.562 & 0.569\\
    GroupTrans~\cite{GroupTrans}  &  0.662 & 0.606 & 0.633 \\
    P-HAR~\cite{DBLP:conf/eccv/HanYLWFW22} & \blue{0.670} & \blue{0.698} & \blue{0.684}\\
    \hline
    Ours  &  \red{0.742} & \red{0.844} & \red{0.790} \\
    
    \hline
    \end{tabular}
   }
\vspace*{4.0mm}
  \centering
  \caption{Quantitative results of static group detection on Caf\'{e}.}
  \label{table: sota_comparison_static_cafe}
    \vspace*{-1.5mm}
   {\footnotesize
   \begin{tabular}{l|ccc}
    \hline
   Method & Precision & Recall & F1 \\
    \hline \hline
    S3R2~\cite{S3R2} & \blue{0.595} & \blue{0.709} & \blue{0.647} \\
    GroupTrans~\cite{GroupTrans}  &  0.286 & 0.425& 0.342 \\
    \hline
    Ours  & \red{0.756} & \red{0.893} & \red{0.819} \\
    \hline
    \end{tabular}
   }
\vspace*{4mm}
\centering
  \caption{Quantitative results of dynamic group detection on JRDB.}
  \label{table: sota_comparison_dynamic_jrdb}
  \vspace*{-1.5mm}
{\footnotesize
\begin{tabular}{l|c|ccc}
\hline
Method &Clustering & Precision & Recall & F1    \\ \hline \hline
\multirow{3}{*}{S3R2~\cite{S3R2}} & LP & 0.522 & 0.662  & 0.584\\
& CNM & 0.568 & 0.360  & 0.440 \\
& Louvain & 0.590 & 0.438 & 0.502 \\ \hline
\multirow{3}{*}{P-HAR~\cite{DBLP:conf/eccv/HanYLWFW22}} & LP & 0.530 & 0.745  & 0.619\\
& CNM & 0.581 & 0.532  & 0.555 \\
& Louvain & 0.568 & 0.697 & 0.626 \\ \hline
\multirow{3}{32pt}{Group Trans~\cite{GroupTrans}} & LP  & 0.514 & 0.534 & 0.524     \\
& CNM & 0.564 & 0.181  & 0.274 \\
& Louvain  & 0.596     & 0.264  & 0.366 \\ \hline
\multirow{3}{*}{Ours} & LP & 0.627& 0.577  & 0.601\\
& CNM & \blue{0.709} & \blue{0.754}  & \blue{0.731} \\
& Louvain & \red{0.724} & \red{0.820} & \red{0.769} \\ \hline
\end{tabular}
}
\vspace*{4.0mm}
  \centering
  \caption{Quantitative results of dynamic group detection on Caf\'{e}.}
  \label{table: sota_comparison_dynamic_cafe}
    \vspace*{-1.5mm}
{\footnotesize
\begin{tabular}{l|c|ccc}
\hline
Method & Clustering & Precision & Recall & F1    \\ \hline \hline
\multirow{3}{*}{S3R2~\cite{S3R2}} & LP & 0.577 & 0.528  & 0.550\\
& CNM & 0.576 & 0.700  & 0.631 \\
& Louvain & 0.572 & 0.633 & 0.600 \\ \hline
\multirow{3}{32pt}{Group Trans~\cite{GroupTrans}} & LP & 0.210 & 0.196 & 0.202     \\
& CNM & 0.087 & 0.069  & 0.077 \\
& Louvain  & 0.263     & 0.305  & 0.283 \\ \hline
\multirow{3}{*}{Ours} & LP & \red{0.739} & 0.768  & 0.753\\
& CNM & 0.648& \red{0.921}  & \blue{0.759} \\
& Louvain & \blue{0.681} & \blue{0.904} & \red{0.776} \\ \hline
\end{tabular}
}
\vspace*{-2mm}
\end{table}

\begin{figure}[t]
  \begin{center}
  \includegraphics[width=\columnwidth]{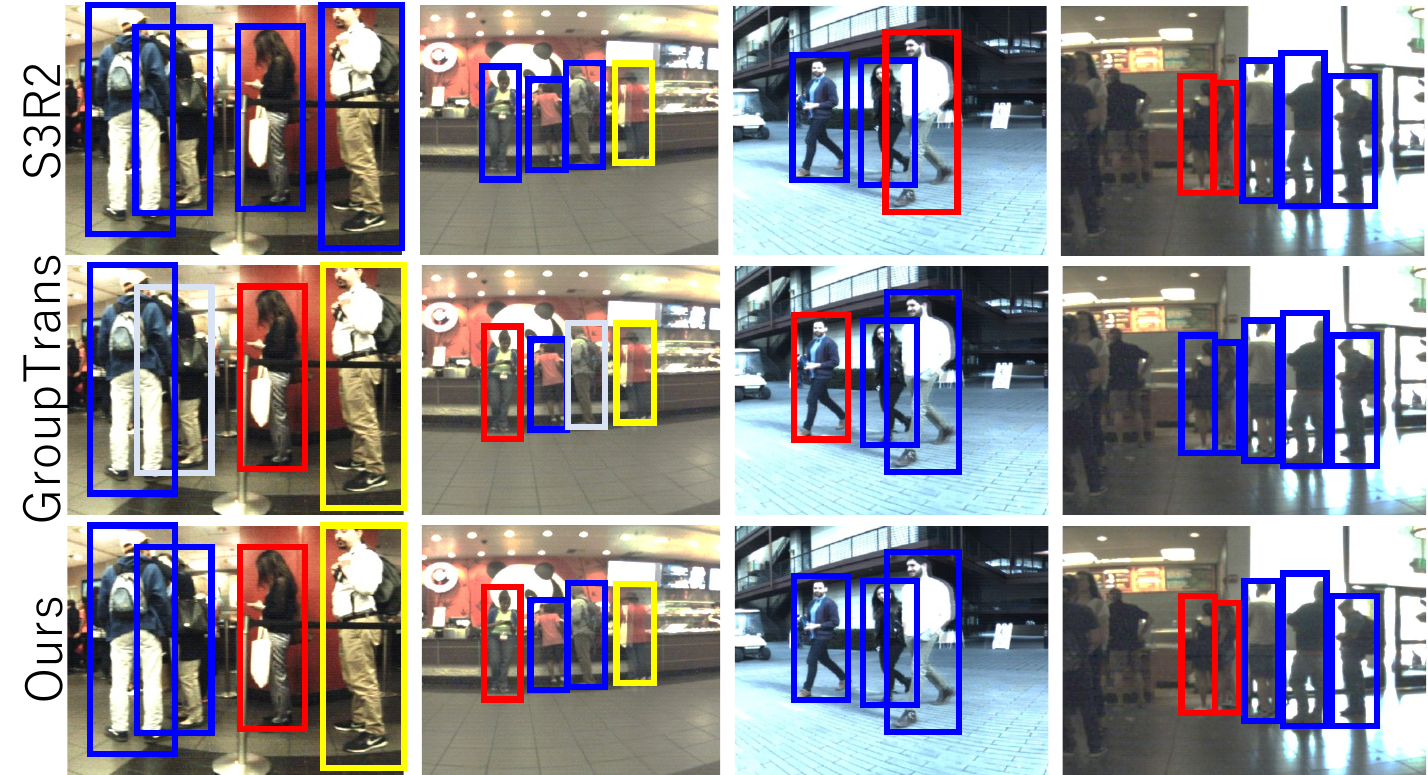}
  \end{center}
  \vspace*{-5mm}
  \hspace*{12mm}
  (a)
  \hspace*{14mm}
  (b)
  \hspace*{14mm}
  (c)
  \hspace*{14mm}
  (d)
  \vspace*{-1mm}
  \caption{Success cases of our method in static group detection on JRDB.
  In Figs.~\ref{fig: static_detection_results} and \ref{fig: dynamic_detection_results}, people enclosed by
  boxes with the same color are detected as in-group members.
  }
  \label{fig: static_detection_results}
  \begin{center}
  \includegraphics[width=\columnwidth]{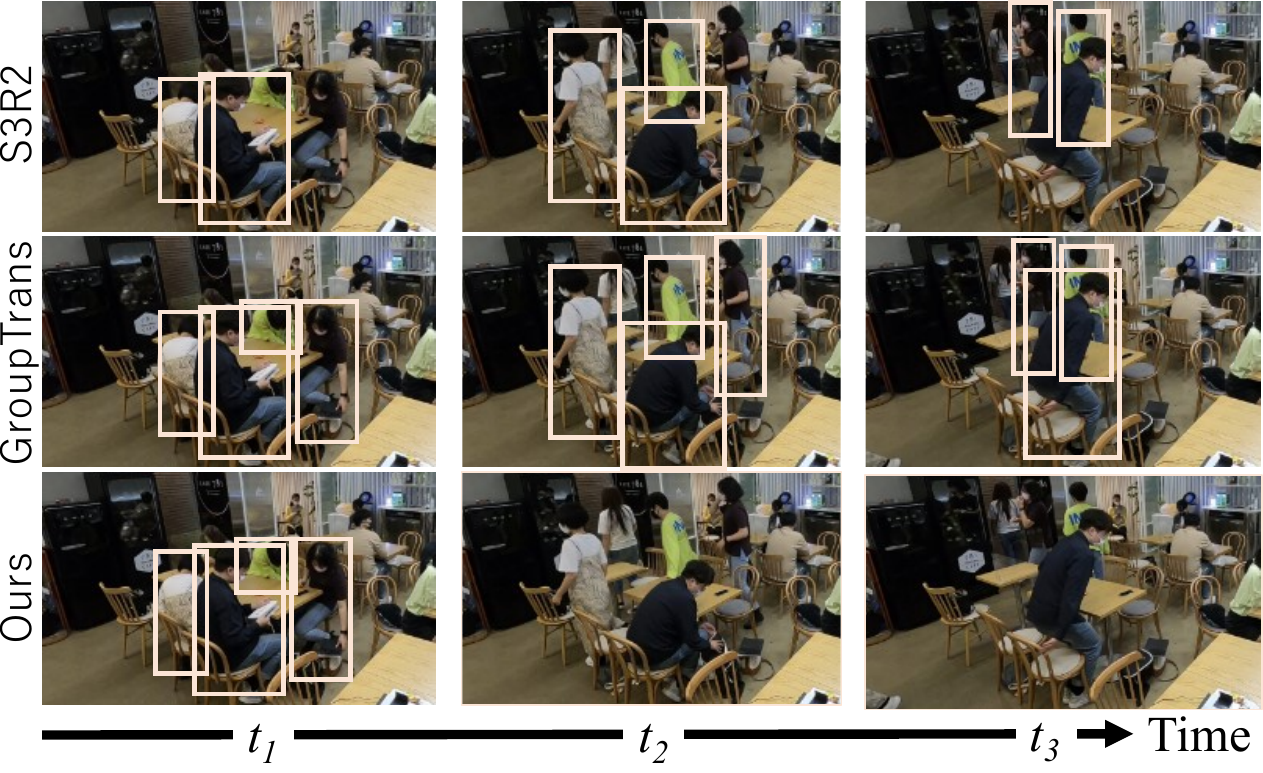}
  \\
  \includegraphics[width=\columnwidth]{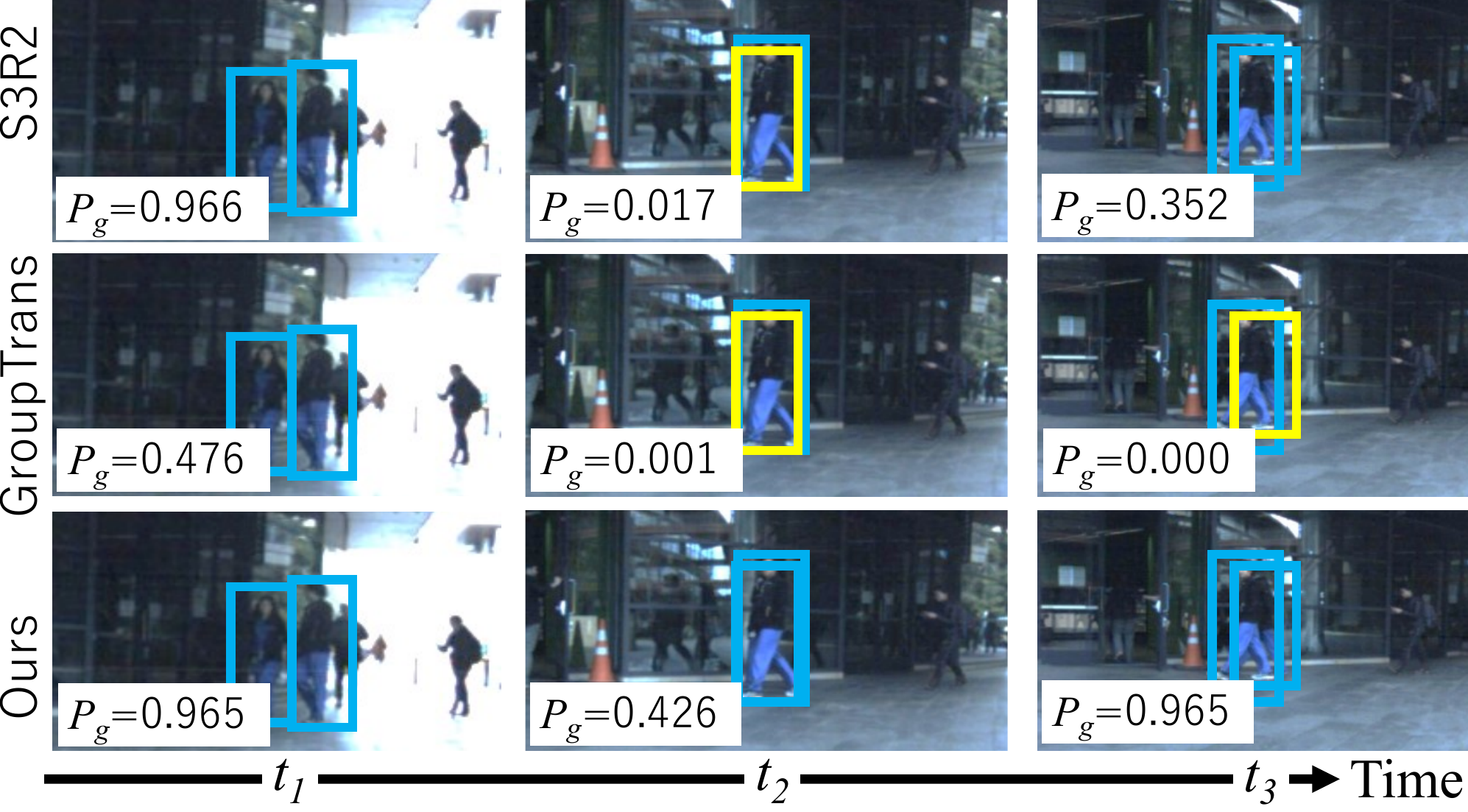}
  \end{center}
  \vspace*{-5mm}
  \caption{Success cases of our method in dynamic group detection on Caf\'{e} (upper) and JRDB (lower).
  In this figure, S3R2 and GroupTransformer are implemented w/ label propagation to contrast with our proposed method using the Louvain algorithm.}
  \label{fig: dynamic_detection_results}
\end{figure}

\noindent
{\bf Static group detection:}
The results
on JRDB and Caf\'{e} are shown in Tables~\ref{table: sota_comparison_static_JRDB} and~\ref{table: sota_comparison_static_cafe}, respectively.
%
The results of all previous methods in Table~\ref{table: sota_comparison_static_JRDB} come from~\cite{S3R2, GroupTrans}, except for P-HAR~\cite{DBLP:conf/eccv/HanYLWFW22} whose result is obtained by the authors' code. In Table~\ref{table: sota_comparison_static_cafe}, the results are obtained by the authors' code.
%
Our method outperforms the previous methods in all metrics on both datasets.
For example, compared with the second-best scores, our method improves F1 by $0.157 = 0.790-0.633$ and $0.172 = 0.819-0.647$ on JRDB and Caf\'{e}, respectively.

Four {\em success} cases of our method in which our GA-CLIP image features are beneficial for group detection are shown in Fig.~\ref{fig: static_detection_results}, which also shows the results of two SoTA methods, S3R2~\cite{S3R2} and GroupTransformer~\cite{GroupTrans}.
In Fig.~\ref{fig: static_detection_results} (a), our method can detect a group in which in-group members are
interacting.
In Fig.~\ref{fig: static_detection_results} (b), our method can discriminate between in-group people and others because the GA-CLIP image features can find the indirect interaction at the counter (i.e., background).
In Fig.~\ref{fig: static_detection_results} (c), the facial directions of people are useful in our method. Figure~\ref{fig: static_detection_results} (d) is a difficult example in which people are just in a queue.
While no clearly distinguishable interaction between these people is observed, their appearance attributes, such as fine-grained face directions, age, and gender, may be used
in our method.

\noindent
{\bf Dynamic group detection:}
The results of dynamic group detection on JRDB and Caf\'{e} are shown in Table~\ref{table: sota_comparison_dynamic_jrdb} and Table~\ref{table: sota_comparison_dynamic_cafe}.
Our method outperforms all others in terms of all metrics by a large gap so that all best and second-best scores are obtained by the variants of our method.
Even compared with the best F1 score among the other methods (i.e., ``0.584 of S3R2 and LP on JRDB'' and ``0.631 of S3R2 and CNM on Caf\'{e}''), our method is better by $0.185 = 0.769 - 0.584$ and $0.145 = 0.776 - 0.631$ on JRDB and Caf\'{e}, respectively.

Two {\em success} cases of our method are shown in Fig.~\ref{fig: dynamic_detection_results}.
In the upper, four sitting people are in the same group at $t_{1}$, while S3R2 fails to include two people to this group.
This group breaks up at $t_{2}$.
Our method detects this breakup soon at $t_{2}$ even though the four people are still close to each other.
This success case validates that (1) our GA-CLIP features can discriminate between in-group people in the table and breaking up people leaving the table, and (2) graph clustering can cut temporal edges
in the temporal groupness graph for dynamic group detection.
In the lower, two people enclosed by blue boxes at $t_{1}$ are mutually occluded
at $t_{2}$.
This occlusion drops their groupness probability, $P_{g}$, at $t_{2}$ compared with those at $t_{1}$ and $t_{3}$.
However, our CLIP finetuning with the occlusion class reduces the drop.
Combining this effect and graph clustering in our temporal groupness graph enables continuous dynamic group detection.


\subsection{Detailed Analyses}
\label{subsection: exp_analysis}

\begin{figure}[t]
  \begin{center}
  \includegraphics[width=\columnwidth]{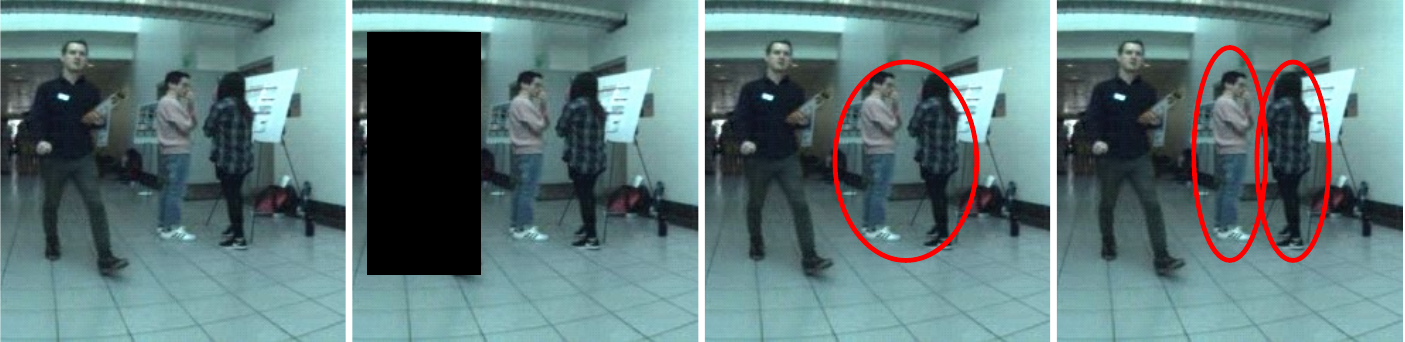}
  \end{center}
  \vspace*{-5mm}
  {\normalsize
  ~\hspace*{0mm}
  (a) No prompt
  ~\hspace*{2mm}
  (b) Mask
  ~\hspace*{2mm}
  (c) A circle
  ~\hspace*{2mm}
  (d) Two circles
  }
  \vspace*{-5mm}
  \caption{Visual prompts for specifying a pair of people of interest.}
  \label{fig: visual_prompts}
\end{figure}

\begin{table}[t]
  \centering
  \caption{Different visual prompts for CLIP.}
  \label{table: visual_prompts}
   {\footnotesize
   \begin{tabular}{l|l|c|ccc}
    \hline
   \multirow{2}{36pt}{Dataset (Task)} & \multirow{2}{*}{Model} & I-Enc & \multicolumn{3}{c}{Group detection} \\
   \cline{3-6} & & F1 & Precision & Recall & F1 \\ \hline \hline
    \multirow{4}{36pt}{JRDB (static)} & No prompt & 0.442 &  0.719 &  0.751 &  0.735\\
    & Mask & 0.569 & 0.653 &  0.823 &  0.728\\
    & A circle & 0.578 & 0.700 &  0.757 &  0.728\\
    & Two circles & \red{0.666} &  \red{0.742} & \red{0.844} & \red{0.790}\\ \hline
    \multirow{4}{36pt}{JRDB (dynamic)} & No prompt & 0.442 & 0.700 & 0.689 & 0.695\\
    & Mask & 0.569  & 0.584  & 0.750 & 0.657\\
    & A circle & 0.578 & 0.640 &  0.666 &  0.653\\
    & Two circles  & \red{0.666} &  \red{0.724} & \red{0.820} & \red{0.769}\\ \hline
    \multirow{4}{36pt}{Caf\'{e} (static)}      & No prompt & 0.580 &  0.591 & 0.774 & 0.670 \\
    & Mask & 0.824 &  0.734 & 0.871 & 0.796 \\
    & A circle & 0.761 &  0.656 & 0.802 & 0.722 \\
    & Two circles & \red{0.885} & \red{0.756} & \red{0.893} & \red{0.819}\\ \hline
    \multirow{4}{36pt}{Caf\'{e} (dynamic)}      & No prompt & 0.580 &  0.508 & 0.777 & 0.615 \\
    & Mask & 0.824 &  0.677 & 0.870 & 0.761 \\
    & A circle & 0.761 &  0.613 & 0.797 & 0.692 \\
    & Two circles & \red{0.885} & \red{0.681} & \red{0.904} & \red{0.776}\\ \hline
    \end{tabular}
   }
\vspace*{4mm}
  \centering
  \caption{Different CLIP finetuning labels, w/ and w/o occlusion.}
  \label{table: ablation_occlusion}
   {\footnotesize
   \begin{tabular}{l|l|ccc}
    \hline
   \multirow{2}{36pt}{Dataset (Task)} & \multirow{2}{*}{Model} & \multirow{2}{*}{Precision} & \multirow{2}{*}{Recall} & \multirow{2}{*}{F1} \\
   &&&&\\
    \hline \hline
    \multirow{2}{36pt}{JRDB (static)} & Ours w/o Occ  &  0.706 & 0.824 & 0.760\\
    & Ours     &  \red{0.742} & \red{0.844} & \red{0.790}\\
    \hline
    \multirow{2}{36pt}{JRDB (dynamic)} & Ours w/o Occ  &  0.637 & 0.743 & 0.686\\
    & Ours     &  \red{0.724} & \red{0.820} & \red{0.769}\\
    \hline
    \end{tabular}
   }
\end{table}


\subsubsection{Different Visual Prompts for CLIP}
\label{subsubsection: exp_prompt}

Table~\ref{table: visual_prompts} shows the contributions of visual prompts, which are shown in Fig.~\ref{fig: visual_prompts}, for group detection.
We can see that our proposed visual prompt using (d) two circles is the best in all datasets and all metrics for group detection.

In addition to group detection, the results of three-class classification (i.e., ``individual people,'' ``a group of people,'' and ``occlusion'') using the image encoder are also shown in the ``I-Enc'' row of Table~\ref{table: visual_prompts}.
The superior performance of (d) two circles validates that the well-trained image encoder contributes to group detection.


\subsubsection{Occlusion Handling with CLIP Finetuning}
\label{subsubsection: exp_occlusion}

The effectiveness of the ``occlusion'' class is verified by ablating it in our CLIP finetuning scheme.
Since ``occlusion'' is not annotated on Caf\'{e}, Table~\ref{table: ablation_occlusion} shows the results on JRDB only.
We can see that ``occlusion'' improves all the metrics in all datasets.
In particular, the improvement of F1 is larger in dynamic group detection (i.e., $0.083 = 0.769 - 0.686$).
This is because occlusion makes group detection difficult in each frame, so the performance is significantly degraded in dynamic detection if the CLIP image encoder does not learn ``occlusion.''
However, by merging the results of dynamic detections for static detection (as described in ``Tasks'' in Sec.~\ref{subsection: exp_dataset}), the results of static detection can be improved, resulting in the small gap between ``Ours w/o Occ'' and ``Ours.''


\begin{table}[t]
\centering
\caption{GA-CLIP image features vs visual features.}
\label{table: ablation_clip}
\vspace*{-2mm}
{\footnotesize
\begin{tabular}{l|l|c|ccc}\hline
&& I-Enc& \multicolumn{3}{c}{Group detection}\\ \cline{3-6}
\multirow{-2}{36pt}{Dataset (Task)} & \multirow{-2}{*}{Model} & F1& Precision& Recall& F1\\ \hline\hline
& ResNet50 & 0.501& \red{0.760}& 0.743& 0.751\\
& ViT-L/16 & 0.523& 0.704& 0.732& 0.718\\
& DINOv2& 0.618& 0.726& 0.746& 0.736\\
\multirow{-4}{36pt}{JRDB (static)}   & Ours &  \red{0.666} & 0.742 & \red{0.844} & \red{0.790} \\ \hline
& ResNet50& 0.501& 0.686& 0.609& 0.645\\
& ViT-L/16& 0.523& 0.652&0.666& 0.659\\
& DINOv2 & 0.618& 0.612&0.600& 0.644\\
\multirow{-4}{36pt}{JRDB (dynamic)}  & Ours &\red{0.666} &\red{0.724} & \red{0.820} & \red{0.769} \\ \hline
& ResNet50& 0.662& 0.599&0.821& 0.692\\
& ViT-L/16& 0.545& 0.538&0.583& 0.554\\
& DINOv2& 0.736& 0.590& 0.759 & 0.664\\
\multirow{-4}{36pt}{Caf\'{e} (static)} & Ours & \red{0.885} & \red{0.756} & \red{0.893} & \red{0.819}\\ \hline 
& ResNet50& 0.662& 0.525&0.806& 0.634\\
& ViT-L/16& 0.545& 0.510&0.579& 0.534\\
& DINOv2& 0.736& 0.555& 0.764 & 0.643\\
\multirow{-4}{36pt}{Caf\'{e} (dynamic)} & Ours  & \red{0.885} & \red{0.681} & \red{0.904} & \red{0.776}\\
\hline
\end{tabular}
}
\vspace*{2mm}
\centering
\caption{Feature ablation study.}
\label{table: ablation_features}
\vspace*{-2mm}
{\footnotesize
\begin{tabular}{l|l|ccc}
\hline
{Dataset (Task)} & {Model} & {Precision} & {Recall} & {F1} \\
\hline \hline
\multirow{3}{36pt}{JRDB (static)} & Ours w/o App.  &  0.636 & 0.771 & 0.697\\
& Ours w/o Traj. &  0.598 & 0.807 & 0.687\\
& Ours     &  \red{0.742} & \red{0.844} & \red{0.790}\\
\hline
\multirow{3}{36pt}{JRDB (dynamic)} & Ours w/o App.  &  0.539 & 0.682 & 0.602\\
& Ours w/o Traj. &  0.534 & 0.748 & 0.623\\
& Ours     &  \red{0.724} & \red{0.820} & \red{0.769}\\
\hline
\multirow{3}{36pt}{Caf\'{e} (static)}      & Ours w/o App.  &  0.640 & 0.837 & 0.724\\
& Ours w/o Traj. &  0.667 & 0.854 & 0.748\\
& Ours     & \red{0.756} & \red{0.893} & \red{0.819}\\
\hline
\multirow{3}{36pt}{Caf\'{e} (dynamic)}      & Ours w/o App.  &  0.555 & 0.848 & 0.670\\
& Ours w/o Traj. &  0.612 & 0.850 & 0.711\\
& Ours     & \red{0.681} & \red{0.904} & \red{0.776} \\
\hline
\end{tabular}
}
\vspace*{-2mm}
\end{table}

\subsubsection{GA-CLIP image features vs Visual Features}
\label{subsubsection: CLIP_vs_visual}

Large pretrained vision encoders, ResNet~\cite{resnet}, ViT~\cite{visiontransformer}, and DINOv2~\cite{dinov2}, are finetuned with each dataset and then replace the CLIP image encoder.
In addition to group detection, three-class classification with the
encoder
is also evaluated, as shown in the ``I-Enc'' row.

Table~\ref{table: ablation_clip} shows that in terms of F1, our method using GA-CLIP outperforms the others in all cases by a large margin.
This proves that higher-level spatial contexts useful for group detection are represented in GA-CLIP.



\subsubsection{Ablation of Trajectory and Image Features}
\label{subsubsection: exp_image_features}

In 
Table~\ref{table: ablation_features}, our method with both appearance and trajectory features is better than that using one of them in all cases.
While there is not much difference between ours w/o appearance features and ours w/o trajectory features on the whole, our method using both features can outperform both ours w/o appearance and ours w/o trajectory.


\subsubsection{Inference Time}
\label{subsubsection: exp_time}

On a NVIDIA RTX 6000 Ada, the image-trajectory fusion features are computed in 0.018 secs/pair.
Since the average number of all possible pairs in each frame
is 958 in JRDB, the runtime is $0.018 \times 958 \risingdotseq 17.2$ secs/frame.
To reduce the runtime, (1) only $k$-nearest neighbor pairs for each person are chosen to reduce the number of pairs to 47.6, as mentioned in the Supp, and (2) 300 pairs are handled in parallel, leading to $0.018 \times \frac{47.6}{300} \risingdotseq 0.003$ secs/frame.
Note that, while only $k$-NN pairs are chosen, temporal graph clustering allows us to detect groups, each of which has $k$ or more people, as validated in the Supp.
The Louvain needs 0.020 secs/frame.
On JRDB, therefore, the total inference time is 0.020 + 0.003 = 0.023 secs/frame, which is faster than 0.11 secs/frame of S3R2~\cite{S3R2}.
On Caf\'{e}, the total time is 0.007 secs/frame, while it is 0.018 secs/frame in S3R2~\cite{S3R2}.


\subsubsection{Human Bounding Boxes given by Tracking}
\label{subsubsection: exp_tracking}

\begin{table}[t]
\centering
  \caption{Dynamic group detection using human bounding boxes given by multi-object tracking~\cite{DBLP:journals/ijcv/ZhangWWZL21} on JRDB.}
  \label{table: mot_group_detection}
  \vspace*{-2mm}
{\footnotesize
\begin{tabular}{l|c|ccc}
\hline
Method & Clustering & Precision & Recall & F1\\ \hline \hline
\multirow{1}{70pt}{S3R2~\cite{S3R2}}& Louvain & 0.617 & 0.725 & 0.667 \\
\multirow{1}{70pt}{Group Trans~\cite{GroupTrans}}&Louvain  & 0.532 & 0.389  & 0.449 \\ \hline
\multirow{1}{70pt}{Ours} & Louvain & \red{0.709} & \red{0.850} & \red{0.774} \\ \hline
\end{tabular}
}
\end{table}

Instead of annotated 
person bounding boxes and IDs between frames, inaccurate 
ones are obtained by a multi-object tracking algorithm~\cite{DBLP:journals/ijcv/ZhangWWZL21}, which needs to be used in real scenarios.
Table~\ref{table: mot_group_detection} shows the superiority of our method.
This may be because our GA-CLIP features play a more crucial role in redeeming the negative effect of erroneous weights given to temporal edges (i.e., erroneous multi-object tracking probabilities).
Even if these temporal edge weights are unreliable, our GA-CLIP features allow for correct framewise group detection.


\section{Limitations}
\label{section: limitation}

\begin{figure}[t]
  \begin{center}
  \includegraphics[width=\columnwidth]{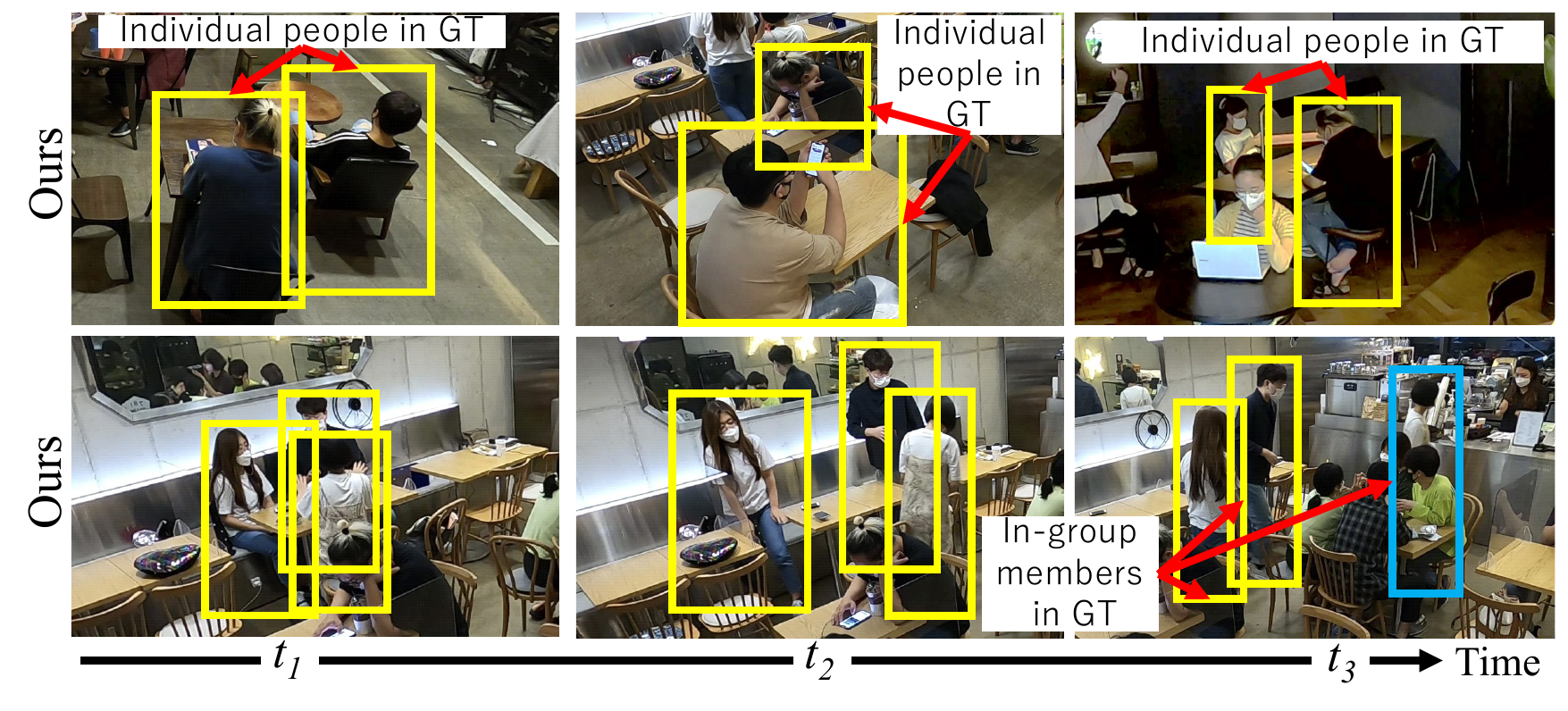}
  \end{center}
  \vspace*{-7mm}
  \caption{Failure cases of our method on Caf\'{e}. Detected in-group members are marked by boxes with the same color.}
  \label{fig: limitation_cafe}
  \vspace*{-2mm}
\end{figure}

Figure~\ref{fig: limitation_cafe} shows failure cases of our method.
In all false-positive group detections shown in the upper, individual people are detected as in-group people because they are nearby for a long time. 
The lower shows three temporal frames.
While three in-group people are detected correctly at $t_{1}$ and $t_{2}$, group detection is failed at $t_{3}$ because one of them enclosed by the blue box walks a different path.
These failure cases reveal that our method stlil depends more on trajectory features than image features, while our proposed GA-CLIP features improve framewise group detection.


\section{Concluding Remarks}
\label{section: conclusion}

We proposed (1) the GA-CLIP image features for framewise groupness evaluation and (2) temporal groupness graph clustering
for dynamic group evaluation.
Our experiments demonstrated that (1) the GA-CLIP image features represent human-human and human-scene interactions for better group detection and (2) temporal groupness graph clustering enables dynamic group detection, which is not achieved even in SoTA methods~\cite{S3R2,GroupTrans}.


{
    \small
    \bibliographystyle{ieeenat_fullname}
    \bibliography{references}
}

\end{document}